\documentclass[twoside,11pt]{article}

\usepackage{styles}
\usepackage{booktabs}
\usepackage{multirow}

\jmlrheading{}{}{}{}{}{Dengliang Shi}

\ShortHeadings{A Study on Neural Network Language Modeling}{D. Shi}
\firstpageno{1}

\begin{document}

\title{A Study on Neural Network Language Modeling}

\author{\name Dengliang Shi \email dengliang.shi@yahoo.com \\
       \addr Shanghai, Shanghai, China}

\editor{}

\maketitle

\begin{abstract}
An exhaustive study on neural network language modeling (NNLM) is performed in this paper. Different architectures of basic neural network language models are described and examined. A number of different improvements over basic neural network language models, including importance sampling, word classes, caching and bidirectional recurrent neural network (BiRNN), are studied separately, and the advantages and disadvantages of every technique are evaluated. Then, the limits of neural network language modeling are explored from the aspects of model architecture and knowledge representation. Part of the statistical information from a word sequence will loss when it is processed word by word in a certain order, and the mechanism of training neural network by updating weight matrixes and vectors imposes severe restrictions on any significant enhancement of NNLM. For knowledge representation, the knowledge represented by neural network language models is the approximate probabilistic distribution of word sequences from a certain training data set rather than the knowledge of a language itself or the information conveyed by word sequences in a natural language. Finally, some directions for improving neural network language modeling further is discussed.
\end{abstract}


\section{Introduction}

Generally, a well-designed language model makes a critical difference in various natural language processing (NLP) tasks, like speech recognition \citep{hinton_2012, graves_2013a}, machine translation \citep{cho_2014a, wu_2016}, semantic extraction \citep{collobert_2007, collobert_2008} and etc. Language modeling (LM), therefore, has been the research focus in NLP field all the time, and a large number of sound research results have been published in the past decades. N-gram based LM \citep{goodman_2001a}, a non-parametric approach, is used to be state of the art, but now a parametric method - neural network language modeling (NNLM) is considered to show better performance and more potential over other LM techniques, and becomes the most commonly used LM technique in multiple NLP tasks.

Although some previous attempts \citep{miikk_1991, schmidhuber_1996, xu_2000} had been made to introduce artificial neural network (ANN) into LM, NNLM began to attract researches' attentions only after \citet{bengio_2003a} and did not show prominent advantages over other techniques of LM until recurrent neural network (RNN) was investigated for NNLM \citep{mikolov_2010, mikolov_2011}. After more than a decade's research, numerous improvements, marginal or critical, over basic NNLM have been proposed. However, the existing experimental results of these techniques are not comparable because they were obtained under different experimental setups and, sometimes, these techniques were evaluated combined with other different techniques. Another significant problem is that most researchers focus on achieving a state of the art language model, but the limits of NNLM are rarely studied. In a few works \citep{jozefowicz_2016} on exploring the limits of NNLM, only some practical issues, like computational complexity, corpus, vocabulary size, and etc., were dealt with, and no attention was spared on the effectiveness of modeling a natural language using NNLM.

Since this study focuses on NNLM itself and does not aim at raising a state of the art language model, the techniques of combining neural network language models with other kind of language models, like N-gram based language models, maximum entropy (ME) language models and etc., will not be included. The rest of this paper is organized as follows: In next section, the basic neural network language models - feed-forward neural network language model (FNNLM), recurrent neural network language model (RNNLM) and long-short term memory (LSTM) RNNLM, will be introduced, including the training and evaluation of these models. In the third section, the details of some important NNLM techniques, including importance sampling, word classes, caching and bidirectional recurrent neural network (BiRNN), will be described, and experiments will be performed on them to examine their advantages and disadvantages separately. The limits of NNLM, mainly about the aspects of model architecture and knowledge representation, will be explored in the fourth section. A further work section will also be given to represent some further researches on NNLM. In last section, a conclusion about the findings in this paper will be made.

\section{Basic Neural Network Language Models}
The goal of statistical language models is to estimate the probability of a word sequence $w_1w_2...w_T$ in a natural language, and the probability can be represented by the production of the conditional probability of every word given all the previous ones:
\[
P(w_{1}^{T})\;=\;\prod^{T}_{t=1}P(w_t{\mid}w_{1}^{t-1})
\]
where, $w_{i}^{j}=w_iw_{i+1}\dots{w_{j-1}w_j}$. This chain rule is established on the assumption that words in a word sequence only statistically depend on their previous context and forms the foundation of all statistical language modeling. NNLM is a kind of statistical language modeling, so it is also termed as neural probabilistic language modeling or neural statistical language modeling. According to the architecture of used ANN, neural network language models can be classified as: FNNLM, RNNLM and LSTM-RNNLM.

\subsection{Feed-forward Neural Network Language Model, FNNLM}
As mentioned above, the objective of FNNLM is to evaluate the conditional probability $P(w_t{\mid}w_{1}^{t-1})$, but feed-forward neural network (FNN) lacks of an effective way to represent history context. Hence, the idea of n-gram based LM is adopted in FNNLM that words in a word sequence more statistically depend on the words closer to them, and only the $n-1$ direct predecessor words are considered when evaluating the conditional probability, this is:
\[
P(w_t{\mid}w_{1}^{t-1})\;\approx\;P(w_t{\mid}w_{t-n+1}^{t-1})
\]

\begin{figure}[!t]
\centering
\includegraphics[width=5in]{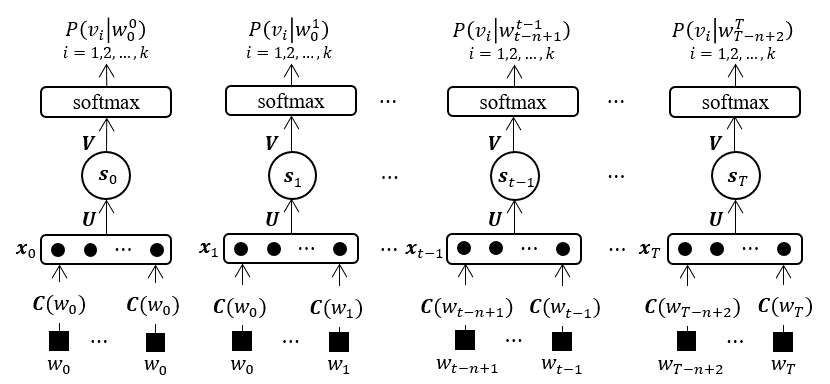}
\caption{Feed-forward neural network language model}
\label{fig:fnnlm}
\end{figure}

The architecture of the original FNNLM proposed by \citet{bengio_2003a} is showed in Figure \ref{fig:fnnlm}, and $w_0$, $w_{T+1}$ are the start and end marks of a word sequence respectively. In this model, a vocabulary is pre-built from a training data set, and every word in this vocabulary is assigned with a unique index. To evaluate the conditional probability of word $w_t$, its $n-1$ direct previous words $w_{t-n+1},...,w_{t-1}$ are projected linearly into feature vectors using a shared matrix $\textbf{\textit{C}}\in{\mathbb{R}^{k\times{m}}}$ according to their index in the vocabulary, where $k$ is the size of the vocabulary and $m$ is the feature vectors' dimension. In fact, every row of projection matrix $\textbf{\textit{C}}$ is a feature vector of a word in the vocabulary. The input $\textbf{\textit{x}}\in\mathbb{R}^{n_i}$ of FNN is formed by concatenating the feature vectors of words $w_{t-n+1},...,w_{t-1}$, where $n_i=m\times(n-1)$ is the size of FNN's input layer. FNN can be generally represented as:
\[
\textbf{\textit{y}}\;=\;\textbf{\textit{V}}\cdot{f(\textbf{\textit{U}}\cdot\textbf{\textit{x}}\;+\textbf{\textit{b}})}\;+\;\textbf{\textit{M}}\cdot\textbf{\textit{x}}\;+\;\textbf{\textit{d}}
\]
Where, $\textbf{\textit{U}}\in\mathbb{R}^{n_h\times{n_i}}$,  $\textbf{\textit{V}}\in\mathbb{R}^{n_o\times{n_h}}$ are weight matrixes, $n_h$ is the size of hidden layer, $n_o=k$ is the size of output layer, weight matrix $\textbf{\textit{M}}\in\mathbb{R}^{n_o\times{n_i}}$ is for the direct connections between input layer and output layer, $\textbf{\textit{b}}\in\mathbb{R}^{n_h}$ and $\textbf{\textit{d}}\in\mathbb{R}^{n_o}$ are vectors for bias terms in hidden layer and output layer respectively, $\textbf{\textit{y}}\in\mathbb{R}^{n_o}$ is output vector, and $f(\cdot)$ is activation function.

The $i$-th element of output vector $\textbf{\textit{y}}$ is the unnormalized conditional probability of the word with index $i$ in the vocabulary. In order to guarantee all the conditional probabilities of words positive and summing to one, a softmax layer is always adopted following the output layer of FNN:
\[
P(v_{i}{\mid}w_{1}^{t-1})\;\approx\;P(v_{i}{\mid}w_{t-n+1}^{t-1})=\;\frac{e^{y(v_{i},w_{t-n+1}^{t-1})}}{\sum_{j=1}^{k}e^{y(v_{j},w_{t-n+1}^{t-1})}}, i = 1, 2, ..., k
\]
where $y(v_i,w_{t-n+1}^{t-1})\;(i = 1, 2, ...,k)$ is the $i$-th element of output vector $\textbf{\textit{y}}$, and $v_i$ is the $i$-th word in the vocabulary.

Training of neural network language models is usually achieved by maximizing the penalized log-likelihood of the training data:
\[
L = \frac{1}{T}\sum_{t=1}^{T}\textrm{log}(P(w_t{\mid}w_{1}^{t-1}; \theta))\;+\;R(\theta)
\]
where, $\theta$ is the set of model's parameters to be trained, $R(\theta)$ is a regularization term.

The recommended learning algorithm for neural network language models is stochastic gradient descent (SGD) method using backpropagation (BP) algorithm. A common choice for the loss function is the cross entroy loss which equals to negative log-likelihood here. The parameters are usually updated as:
\[
\theta = \theta + \alpha\frac{\partial{L}}{\partial{\theta}} - \beta\theta
\]
where, $\alpha$ is learning rate and $\beta$ is regularization parameter.

The performance of neural network language models is usually measured using perplexity (PPL) which can be defined as:
\[
PPL\;=\;\sqrt[T]{\prod^{T}_{i=1}\frac{1}{P(w_{i}{\mid}w_{1}^{i-1})}}\;=\;2^{-\frac{1}{T}\sum^{T}_{i=1}log_2P(w_{i}{\mid}w_{1}^{i-1})}
\]
Perplexity can be defined as the exponential of the average number of bits required to encode the test data using a language model and lower perplexity indicates that the language model is closer to the true model which generates the test data.

\subsection{Recurrent Neural Network Language Model, RNNLM}
\begin{figure}[!t]
\centering
\includegraphics[width=4.5in]{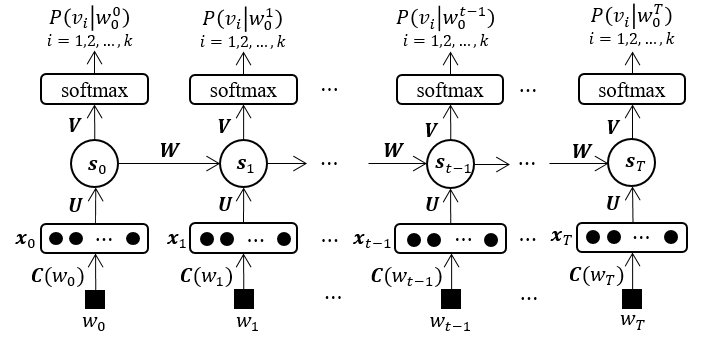}
\caption{Recurrent neural network language model}
\label{fig:rnnlm}
\end{figure}

The idea of applying RNN in LM was proposed much earlier \citep{bengio_2003a, castro_2003}, but the first serious attempt to build a RNNLM was made by \citet{mikolov_2010, mikolov_2011}. RNNs are fundamentally different from feed-forward architectures in the sense that they operate on not only an input space but also an internal state space, and the state space enables the representation of sequentially extended dependencies. Therefore, arbitrary length of word sequence can be dealt with using RNNLM, and all previous context can be taken into account when predicting next word. As showed in Figure \ref{fig:rnnlm}, the representation of words in RNNLM is the same as that of FNNLM, but the input of RNN at every step is the feature vector of a direct previous word instead of the concatenation of the $n-1$ previous words' feature vectors and all other previous words are taken into account by the internal state of previous step. At step $t$, RNN can be described as:
\begin{eqnarray}
\textbf{\textit{s}}_t&=&f(\textbf{\textit{U}}\cdot\textbf{\textit{x}}_t\;+\;\textbf{\textit{W}}\cdot\textbf{\textit{s}}_{t-1}\;+\;\textbf{\textit{b}}),\nonumber\\
\textbf{\textit{y}}_t&=&\textbf{\textit{V}}\cdot\textbf{\textit{s}}_t\;+\;\textbf{\textit{M}}\cdot\textbf{\textit{x}}_t\;+\;\textbf{\textit{d}}\nonumber
\end{eqnarray}
where, weight matrix $\textbf{\textit{W}}\in\mathbb{R}^{n_h\times{n_h}}$, and the input layer's size of RNN $n_i=m$. The outputs of RNN are also unnormalized probabilities and should be regularized using a softmax layer.

Because of the involvement of previous internal state at every step, back-propagation through time (BPTT) algorithm \citep{rumelhart_1986} is preferred for better performance when training RNNLMs. If data set is treated as a single long word sequence, truncated BPTT should be used and back-propagating error gradient through 5 steps is enough, at least for small corpus \citep{mikolov_2012}. In this paper, neural network language models will all be trained on data set sentence by sentence, and the error gradient will be back-propagated trough every whole sentence without any truncation.

\subsection{Long Short Term Memory RNNLM, LSTM-RNNLM}
Although RNNLM can take all predecessor words into account when predicting next word in a word sequence, but it is quite difficult to be trained over long term dependencies because of the vanishing or exploring problem \citep{hochreiter_1997}. LSTM-RNN was designed aiming at solving this problem, and better performance can be expected by replacing RNN with LSTM-RNN. LSTM-RNNLM was first proposed by \citet{sunder_2012}, and the whole architecture is almost the same as RNNLM except the part of neural network. LSTM-RNN was proposed by \citet{hochreiter_1997} and was refined and popularized in following works \citep{gers_2000, cho_2014b}. The general architecture of LSTM-RNN is:
\begin{eqnarray}
\textbf{\textit{i}}_t&=&\sigma(\textbf{\textit{U}}_i\cdot\textbf{\textit{x}}_t\;+\;\textbf{\textit{W}}_i\cdot\textbf{\textit{s}}_{t-1}\;+\;\textbf{\textit{V}}_i\cdot\textbf{\textit{c}}_{t-1}\;+\;\textbf{\textit{b}}_i)\nonumber\\
\textbf{\textit{f}}_t&=&\sigma(\textbf{\textit{U}}_f\cdot\textbf{\textit{x}}_t\;+\;\textbf{\textit{W}}_f\cdot\textbf{\textit{s}}_{t-1}\;+\;\textbf{\textit{V}}_f\cdot\textbf{\textit{c}}_{t-1}\;+\;\textbf{\textit{b}}_f)\nonumber\\
\textbf{\textit{g}}_t&=&f(\textbf{\textit{U}}\cdot\textbf{\textit{x}}_t\;+\;\textbf{\textit{W}}\cdot\textbf{\textit{s}}_{t-1}\;+\;\textbf{\textit{V}}\cdot\textbf{\textit{c}}_{t-1}\;+\;\textbf{\textit{b}})\nonumber\\
\textbf{\textit{c}}_t&=&\textbf{\textit{f}}_t\;*\;\textbf{\textit{c}}_{t-1}\;+\;\textbf{\textit{i}}_t\;*\;\textbf{\textit{g}}_t\nonumber\\
\textbf{\textit{o}}_t&=&\sigma(\textbf{\textit{U}}_o\cdot\textbf{\textit{x}}_t\;+\;\textbf{\textit{W}}_o\cdot\textbf{\textit{s}}_{t-1}\;+\;\textbf{\textit{V}}_o\cdot\textbf{\textit{c}}_{t}\;+\;\textbf{\textit{b}}_o)\nonumber\\
\textbf{\textit{s}}_t&=&\textbf{\textit{o}}_t\;*\;f(\textbf{\textit{c}}_t)\nonumber\\
\textbf{\textit{y}}_t&=&\textbf{\textit{V}}\cdot\textbf{\textit{s}}_t + \textbf{\textit{M}}\cdot\textbf{\textit{x}}_t + \textbf{\textit{d}}\nonumber
\end{eqnarray}
Where, $\textbf{\textit{i}}_t$, $\textbf{\textit{f}}_t$, $\textbf{o}_t\in\mathbb{R}^{n_h}$ are input gate, forget gate and output gate, respectively. $\textbf{c}_t\in\mathbb{R}^{n_h}$ is the internal memory of unit. $\textbf{\textit{U}}_i$, $\textbf{\textit{U}}_f$, $\textbf{\textit{U}}_o$, $\textbf{\textit{U}}\in\mathbb{R}^{n_h\times{n_i}}$, $\textbf{\textit{W}}_i$, $\textbf{\textit{W}}_f$, $\textbf{\textit{W}}_o$, $\textbf{\textit{W}}\in\mathbb{R}^{n_h\times{n_h}}$, $\textbf{\textit{V}}_i$, $\textbf{\textit{V}}_f$, $\textbf{\textit{V}}_o$, $\textbf{\textit{V}}\in\mathbb{R}^{n_h\times{n_h}}$ are all weight matrixes. $\textbf{\textit{b}}_i$, $\textbf{\textit{b}}_f$, $\textbf{\textit{b}}_o$, $\textbf{\textit{b}}\in\mathbb{R}^{n_h}$, and $\textbf{\textit{d}}\in\mathbb{R}^{n_o}$ are vectors for bias terms. $f(\cdot)$ is the activation function in hidden layer and $\sigma(\cdot)$ is the activation function for gates.

\subsection{Comparison of Neural Network Language Models}
Comparisons among neural network language models with different architectures have already been made on both small and large corpus \citep{mikolov_2012, sunder_2013}. The results show that, generally, RNNLMs outperform FNNLMs and the best performance is achieved using LSTM-NNLMs. However, the neural network language models used in these comparisons are optimized using various techniques, and even combined with other kind of language models, let alone the different experimental setups and implementation details, which make the comparison results fail to illustrate the fundamental discrepancy in the performance of neural network language models with different architecture and cannot be taken as baseline for the studies in this paper.

Comparative experiments on neural network language models with different architecture were repeated here. The models in these experiments were all implemented plainly, and only a class-based speed-up technique was used which will be introduced later. Experiments were performed on the Brown Corpus, and the experimental setup for Brown corpus is the same as that in \citep{bengio_2003a}, the first 800000 words (ca01$\sim$cj54) were used for training, the following 200000 words (cj55$\sim$cm06) for validation and the rest (cn01$\sim$cr09) for test.

\begin{table}[!hbp]
\begin{center}
\begin{tabular}{c|c|c|c|c|c|c}
\toprule
Models & $n$ & $m$ & $n_h$ & Direct & Bias & PPL\\
\midrule
FNNLM & 5 & 100 & 200 & No & No & 223.85 \\
RNNLM & - & 100 & 200 & No & No & 221.10 \\
LSTM-RNNLM & - & 100 & 200 & No & No & 237.93 \\
LSTM-RNNLM & - & 100 & 200 & Yes & No & 242.54 \\
LSTM-RNNLM & - & 100 & 200 & No & Yes & 237.18 \\
\bottomrule
\end{tabular}
\caption{Camparative results of different neural network language models}
\label{tab:baseline}
\end{center}
\end{table}

The experiment results are showed in Table \ref{tab:baseline} which suggest that, on a small corpus likes the Brown Corpus, RNNLM and LSTM-RNN did not show a remarkable advantage over FNNLM, instead a bit higher perplexity was achieved by LSTM-RNNLM. Maybe more data is needed to train RNNLM and LSTM-RNNLM because longer dependencies are taken into account by RNNLM and LSTM-RNNLM when predicting next word. LSTM-RNNLM with bias terms or direct connections was also evaluated here. When the direct connections between input layer and output layer of LSTM-RNN are enabled, a slightly higher perplexity but shorter training time were obtained. An explanation given for this phenomenon by \citet{bengio_2003a} is that direct connections provide a bit more capacity and faster learning of the "linear" part of mapping from inputs to outputs but impose a negative effect on generalization. For bias terms, no significant improvement on performance was gained by adding bias terms which was also observed on RNNLM by \citet{mikolov_2012}. In the rest of this paper, all studies will be performed on LSTM-RNNLM with neither direct connections nor bias terms, and the result of this model in Table \ref{tab:baseline} will be used as the baseline for the rest studies.

\section{Optimization Techniques}
\subsection{Importance Sampling}
Inspired by the contrastive divergence model \citep{hinton_2002}, \citet{bengio_2003b} proposed a sampling-based method to speed up the training of neural network language models. In order to apply this method, the outputs of neural network should be normalized in following way instead of using a softmax function:
\[
P(v_i{\mid}w_{1}^{t-1}) = \frac{e^{-y(v_{i},w_{1}^{t-1})}}{\sum_{j=1}^{k}{e^{-y(v_{j},w_{1}^{t-1})}}},\;i=1,2,\dots,k;\;t=1,2,\dots,T
\]
then, neural network language models can be treated as a special case of energy-based probability models.

The main idea of sampling based method is to approximate the average of log-likelihood gradient with respect to the parameters $\theta$ by samples rather than computing the gradient explicitly. The log-likelihood gradient for the parameters set $\theta$ can be generally represented as the sum of two parts: positive reinforcement for target word $w_t$ and negative reinforcement for all word $v_i$, weighted by $P(v_i{\mid}w_{1}^{t-1})$:
\[
\frac{\partial{\textrm{log}P(w_t{\mid}w_{1}^{t-1})}}{\partial{\theta}}\;=\;-\frac{\partial{y(w_t,w_{1}^{t-1})}}{\partial{\theta}}\;+\;\sum_{i=1}^{k}P(v_i{\mid}w_{1}^{t-1})\frac{\partial{y(v_i,w_{1}^{t-1})}}{\partial{\theta}}
\]

Three sampling approximation algorithms were presented by \citet{bengio_2003b}: Monte-Carlo Algorithm, Independent Metropolis-Hastings Algorithm and Importance Sampling Algorithm. However, only importance sampling worked well with neural network language model. In fact, 19-fold speed-up was achieved during training while no degradation of the perplexities was observed on both training and test data \citep{bengio_2003b}.

Importance sampling is a Monte-Carlo scheme using an existing proposal distribution, and its estimator can be represented as:
\[
E[\sum_{x{\sim}P}P(x)g(x)]\;=\;\frac{1}{N}\sum_{x^{'}\in{\Gamma}}g(x^{'})\frac{P(x^{'})}{Q(x^{'})}
\]
where, $Q$ is an existing proposal distribution, $N$ is the number of samples from $Q$, $\Gamma$ is the set of samples from $Q$. Appling importance sampling to the average log-likelihood gradient of negative samples and the denominator of $P(v_{i}{\mid}w_{1}^{t-1})$, then the overall estimator for example $(w_t, w_{1}^{t-1})$ using $N$ samples from distribution $Q$ is:
\[
E[\frac{\partial{\textrm{log}P(w_{t}{\mid}w_{1}^{t-1})}}{\partial{\theta}}]\;=\;-\frac{\partial{y(w_{t}, w_{1}^{t-1})}}{\partial{\theta}}\;+\;\frac{\sum_{w^{'}\in{\Gamma}}\frac{\partial{y(w^{'}{\mid}w_{1}^{t-1})}}{\partial{\theta}}e^{-y(w^{'},w_{1}^{t-1})}/Q(w^{'}{\mid}w_{1}^{t-1})}{\sum_{w^{'}\in{\Gamma}}e^{-y(w^{'},w_{1}^{t-1})}/Q(w^{'}{\mid}w_{1}^{t-1})}
\]

In order to avoid divergence, the sample size $N$ should be increased as training processes which is measured by the effective sample size of importance sampling:
\[
S = \frac{(\sum_{j=1}^{N}r_j)^2}{\sum_{j=1}^{N}r_j^2},\;r_j\;\approx\;\frac{e^{-y(w_{j}^{'},w_{1}^{t-1})}/Q(w_{j}^{'}{\mid}w_{1}^{t-1})}{\sum_{w^{'}\in{\Gamma}}e^{-y(w^{'},w_{1}^{t-1})}/Q(w^{'}{\mid}w_{1}^{t-1})}
\]
At every iteration, sampling is done block by block with a constant size until the effective sample size $S$ becomes greater than a minimum value, and a full back-propagation will be performed when the sampling size $N$ is greater than a certain threshold.

The introduction of importance sampling is just posted here for completeness and no further studies will be performed on it. Because a quick statistical language model which is well trained, like n-gram based language model, is needed to implement importance sampling. In addition, it cannot be applied into RNNLM or LSTM-RNNLM directly and other simpler and more efficient speed-up techniques have been proposed now.

\subsection{Word Classes}
\begin{figure}[!t]
\centering
\includegraphics[width=2in]{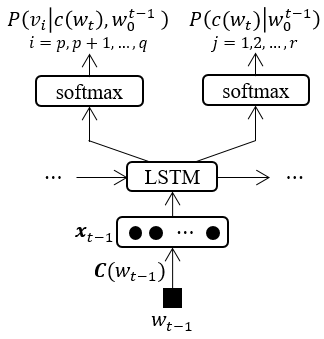}
\caption{Architecture of class based LSTM-RNNLM}
\label{fig:rnnlm-class}
\end{figure}

Before the idea of word classes was introduced to NNLM, it had been used in LM extensively for improving perplexities or increasing speed \citep{brown_1992, goodman_2001b}. With word classes, every word in vocabulary is assigned to a unique class, and the conditional probability of a word given its history can be decomposed into the probability of the word's class given its history and the probability of the word given its class and history, this is:
\[
P(w_{t}{\mid}w_{1}^{t-1})\;=\;P(w_t{\mid}c(w_t),w_{1}^{t-1})P(c(w_t){\mid}w_{1}^{t-1})
\]
where $c(w_t)$ is the class of word $w_t$. The architecture of class based LSTM-RNNLM is illustrated in Figure \ref{fig:rnnlm-class}, and $p$, $q$ are the lower and upper index of words in a class respectively.

\citet{morin_2005} extended word classes to a hierarchical binary clustering of words and built a hierarchical neural network language model. In hierarchical neural network language model, instead of assigning every word in vocabulary with a unique class, a hierarchical binary tree of words is built according to the word similarity information extracted from WordNet \citep{fellbaum_1998}, and every word in vocabulary is assigned with a bit vector $b=[b_{1}(v_i),b_{2}(v_i),\dots,b_{l}(v_i)]$, $i=1,2,\dots,k$. When $b_{1}(v_i),b_{2}(v_i),\dots,b_{j-1}(v_i)$ are given, $b_j(v_i)=0, j=1,2,\dots,l$ indicates that word $v_i$ belongs to the sub-group 0 of current node and $b_j(v_i)=1$ indicates it belongs to the other one. The conditional probability of every word is represented as:
\[
P(v_{i}{\mid}w_{1}^{t-1})\;=\;\prod_{j=1}^{l}P(b_{j}(w_t){\mid}b_{1}(w_t),\dots,b_{j-1}(w_t), w_{1}^{t-1})
\]

Theoretically, an exponential speed-up, on the order of $k/\textrm{log}_{2}k$, can be achieved with this hierarchical architecture. In \citet{morin_2005}, impressive speed-up during both training and test, which were less than the theoretical one, were obtained but an obvious increase in PPL was also observed. One possible explanation for this phenomenon is that the introduction of hierarchical architecture or word classes impose negative influence on the word classification by neural network language models. As is well known, a distribution representation for words, which can be used to represent the similarities between words, is formed by neural network language models during training. When words are clustered into classes, the similarities between words from different classes cannot be recognized directly. For a hierarchical clustering of words, words are clustered more finely which might lead to worse performance, i.e., higher perplexity, and deeper the hierarchical architecture is, worse the performance would be.

To explore this point further, hierarchical LSTM-NNLMs with different number of hierarchical layers were built. In these hierarchical LSTM-NNLMs, words were clustered randomly and uniformly instead of according to any word similarity information. The results of experiment on these models are showed in Table \ref{tab:layers} which strongly support the above hypothesis. When words are clustered into hierarchical word classes, the speed of both training and test increase, but the effect of speed-up decreases and the performance declines dramatically as the number of hierarchical layers increases. Lower perplexity can be expected if some similarity information of words is used when clustering words into classes. However, because of the ambiguity of words, the degradation of performance is unavoidable by assigning every word with a unique class or path. On the other hand, the similarities among words recognized by neural network is hard to defined, but it is sure that they are not confined to linguistical ones.

\begin{table}[!hbp]
\begin{center}
\begin{tabular}{c|c|c|c|c|c|c|c}
\toprule
\multirow{2}{*}{Model} & \multirow{2}{*}{$m$} & \multirow{2}{*}{$n_h$} & \multirow{2}{*}{Method} & \multirow{2}{*}{$l$} & \multirow{2}{*}{PPL} & \multicolumn{2}{c}{Words/s}\\
\cline{7-8}
 & & & & & & Train & Test\\
\midrule
LSTM-NNLM & 100 & 200 & Uniform & 1 & 227.51 & 607.09 & 2798.97 \\
LSTM-NNLM & 100 & 200 & Uniform & 3 & 312.82 & 683.04 & 3704.28 \\
LSTM-NNLM & 100 & 200 & Uniform & 5 & 438.58 & 694.43 & 3520.27 \\
LSTM-NNLM & 100 & 200 & Freq & 1 & 248.99 & 600.56 & 2507.97 \\
LSTM-NNLM & 100 & 200 & Sqrt-Freq & 1 & 237.93 & 650.16 & 3057.27 \\
\bottomrule
\end{tabular}
\caption{Results for class-based models}
\label{tab:layers}
\end{center}
\end{table}

There is a simpler way to speed up neural network language models using word classes which was proposed by \citet{mikolov_2011}. Words in vocabulary are arranged in descent order according to their frequencies in training data set, and are assigned to classes one by one using following rule:
\[
\frac{i}{r}<\sum_{j=1}^{z}\frac{f_j}{F}\leq\frac{i+1}{r},\;0\leq{i}\leq{r-1}
\]
where, $r$ is the target number of word classes, $f_j$ is the frequency of the $j$-th word in vocabulary, the sum of all words' frequencies $F=\sum_{j=1}^{k}f_j$. If the above rule is satisfied, the $z$-th word in vocabulary will be assigned to $i$-th class. In this way, the word classes are not uniform, and the first classes hold less words with high frequency and the last ones contain more low-frequency words. This strategy was further optimized by \citep{mikolov_2012} using following criterion:
\[
\frac{i}{r}<\sum_{j=1}^{z}\frac{\sqrt{f_j/F}}{dF}\leq\frac{i+1}{r}
\]
where, the sum of all words' sqrt frequencies $dF=\sum_{j=1}^{k}\sqrt{f_j/F}$.

The experiment results (Table \ref{tab:layers}) indicate that higher perplexity and a little more training time were obtained when the words in vocabulary were classified according to their frequencies than classified randomly and uniformly. When words are clustered into word classed using their frequency, words with high frequency, which contribute more to final perplexity, are clustered into very small word classes, and this leads to higher perplexity. On the other hand, word classes consist of words with low frequency are much bigger which causes more training time. However, as the experiment results show, both perplexity and training time were improved when words were classified according to their sqrt frequency, because word classes were more uniform when built in this way. All other models in this paper were speeded up using word classes, and words were clustered according to their sqrt frequencies.

\subsection{Caching}
Like word classes, caching is also a common used optimization technique in LM. The cache language models are based on the assumption that the word in recent history are more likely to appear again. In cache language model, the conditional probability of a word is calculated by interpolating the output of standard language model and the probability evaluated by caching, like:
\[
P(w_t{\mid}w_{0}^{t-1})\;=\;\lambda{P_{o}(w_t{\mid}w_{0}^{t-1})}\;+\;(1-\lambda)P_{c}(w_t{\mid}w_{0}^{t-1})
\]
where, $P_{o}(w_t{\mid}w_{0}^{t-1})$ is the output of standard language model, $P_{c}(w_t{\mid}w_{0}^{t-1})$ is the probability evaluated using caching, and $\lambda$ is a constant, $0\leq{\lambda}\leq{1}$.

\citet{soutner_2012} combined FNNLM with cache model to enhance the performance of FNNLM in speech recognition, and the cache model was formed based on the previous context as following:
\[
P_{c}(w_t{\mid}w_{t-N}^{t-1})\;=\;\frac{1}{N}\sum_{j=1}^{N}\rho\delta(w_t, w_{t-j})
\]
where, $\delta(\cdot)$ means Kronecker delta, $N$ is the cache length, i.e., the number of previous words taken as cache, $\rho$ is a coefficient depends on $j$ which is the distance between previous word and target word. A cache model with forgetting can be obtained by lowering $\rho$ linearly or exponentially respecting to $j$. A class cache model was also proposed by \citet{soutner_2012} for the case in which words are clustered into word classes. In class cache model, the probability of target word given the last recent word classes is determined. However, both word based cache model and class one can be defined as a kind of unigram language model built from previous context, and this caching technique is an approach to combine neural network language model with a unigram model.

Another type of caching has been proposed as a speed-up technique for RNNLMs \citep{bengio_2001,kombrink_2011, si_2013, huang_2014}. The main idea of this approach is to store the outputs and states of language models for future prediction given the same contextual history. In \citet{huang_2014}, four caches were proposed, and they were all achieved by hash lookup tables to store key and value pairs: probability $P(w_t{\mid}w_{0}^{t-1})$ and word sequence $w_{0}^{t}$; history $w_{0}^{t-1}$ and its corresponding hidden state vector; history $w_{0}^{t-1}$ and the denominator of the softmax function for classes; history $w_{0}^{t-1}$, class index $c(w_t)$ and the denominator of the softmax function for words. In \citet{huang_2014}, around 50-fold speed-up was reported with this caching technique in speech recognition but, unfortunately, it only works for prediction and cannot be applied during training.

Inspired by the first caching technique, if the previous context can be taken into account through the internal states of RNN, the perplexity is expected to decrease. In this paper, all language models are trained sentence by sentence, and the initial states of RNN are initialized using a constant vector. This caching technique can be implemented by simply initializing the initial states using the last states of direct previous sentence in the same article. However, the experiment result (Table \ref{tab:caching}) shows this caching technique did not work as excepted and the perplexity even increased slightly. Maybe, the Brown Corpus is too small and more data is needed to evaluated this caching technique, as more context is taken into account with this caching technique.

\begin{table}[!hbp]
\begin{center}
\begin{tabular}{c|c|c|c|c}
\toprule
Models & $m$ & $n_h$ & Caching & PPL \\
\midrule
LSTM-NNLM & 100 & 200 & No & 237.93 \\
LSTM-NNLM & 100 & 200 & Yes & 241.45 \\
\bottomrule
\end{tabular}
\caption{Results of Cached LSTM-RNNLM}
\label{tab:caching}
\end{center}
\end{table}

\subsection{Bidirectional Recurrent Neural Network}
In \citet{sutskever_2014}, significant improvement on neural machine translation (NMT) for an English to French translation task was achieved by reversing the order of input word sequence, and the possible explanation given for this phenomenon was that smaller "minimal time lag" was obtained in this way. In my opinion, another possible explanation is that a word in word sequence may more statistically depend on the following context than previous one. After all, a number of words are determined by its following words instead of previous ones in some natural languages. Take the articles in English as examples, indefinite article "an" is used when the first syllable of next word is a vowel while "a" is preposed before words starting with consonant. What's more, if a noun is qualified by an attributive clause, definite article "the" should be used before the noun. These examples illustrate that words in a word sequence depends on their following words sometimes. To verify this hypothesis further, an experiment is performed here in which the word order of every input sentence is reversed, and the probability of word sequence $w_1w_2{\dots}w_T$ is evaluated as following:
\[
P(w_{1}^{T})\;=\;\prod^{T}_{t=1}P(w_t{\mid}w_{t+1}^{T})
\]

However, the experiment result (Table \ref{tab:birnn}) shows that almost the same perplexity was achieved by reversing the order of words. This indicates that the same amount statistical information, but not exactly the same statistical information, for a word in a word sequence can be obtained from its following context as from its previous context, at least for English. 

\begin{table}[!hbp]
\begin{center}
\begin{tabular}{l|c|c|c|c}
\toprule
Model & $m$ & $n_h$ & Reverse & PPL \\
\midrule
LSTM-NNLM & 100 & 200 & No & 237.93 \\
LSTM-NNLM & 100 & 200 & Yes & 240.48 \\
\bottomrule
\end{tabular}
\caption{Reverse the order of word sequence}
\label{tab:birnn}
\end{center}
\end{table}

\begin{figure}[!t]
\centering
\includegraphics[width=4.5in]{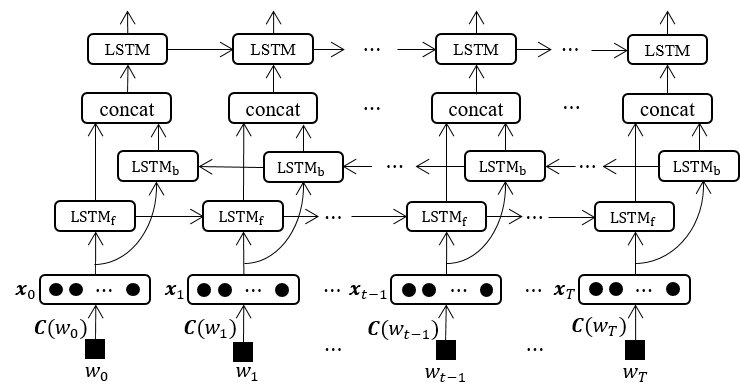}
\caption{Encode word sequence using bidirectional recurrent neural network}
\label{fig:bidirectional}
\end{figure}

As a word in word sequence statistically depends on its both previous and following context, it is better to predict a word using context from its both side. Bidirectional recurrent neural network (BiRNN) \citep{schuster_1997} was designed to process data in both directions with two separate hidden layers, so better performance can be expected by using BiRNN. BiRNN was introduced to speech recognition by \citet{graves_2013b}, and then was evaluated in other NLP tasks, like NMT \citep{bahdanau_2015, wu_2016}. In these studies, BiRNN showed more excellent performance than unidirectional RNN. Nevertheless, BiRNN cannot be evaluated in LM directly as unidirectional RNN, because statistical language modeling is based on the chain rule which assumes that word in a word sequence only statistically depends on one side context. BiRNN can be applied in NLP tasks, like speech recognition and machine translation, because the input word sequences in these tasks are treated as a whole and usually encoded as a single vector. The architecture for encoding input word sequences using BiRNN is showed in Figure \ref{fig:bidirectional}. The facts that better performance can be achieved using BiRNN in speech recognition or machine translation indicate that a word in a word sequence is statistically determined by the words of its both side, and it is not a suitable way to deal with word sequence in a natural language word by word in an order.

\section{Limits of Neural Network Language Modeling}
NNLM is state of the art, and has been introduced as a promising approach to various NLP tasks. Numerous researchers from different areas of NLP attempt to improve NNLM, expecting to get better performance in their areas, like lower perplexity on test data, less word error rate (WER) in speech recognition, higher Bilingual Evaluation Understudy (BLEU) score in machine translation and etc. However, few of them spares attention on the limits of NNLM. Without a thorough understanding of NNLM's limits, the applicable scope of NNLM and directions for improving NNLM in different NLP tasks cannot be defined clearly. In this section, the limits of NNLM will be studied from two aspects: model architecture and knowledge representation.

\begin{figure}[!t]
\centering
\includegraphics[width=5.5in]{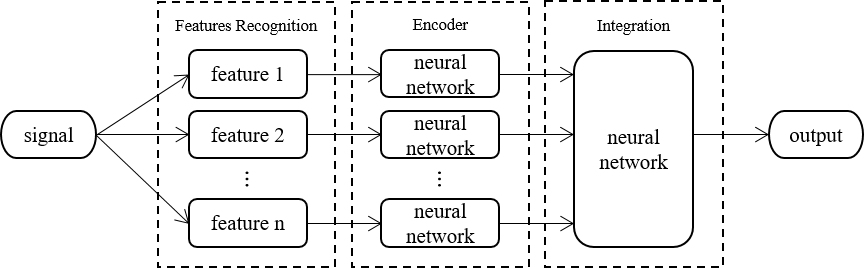}
\caption{A possible scheme for the architecture of ANN}
\label{fig:scheme}
\end{figure}

\subsection{Model Architecture}
In most language models including neural network language models, words are predicated one by one according to their previous context or following one which is believed to simulate the way human deal with natural languages, and, according to common sense, human actually speak or write word by word in a certain order. However, the intrinsic mechanism in human mind of processing natural languages cannot like this way. As mentioned above, it is not always true that words in a word sequence only depend on their previous or following context. In fact, before human speaking or writing, they know what they want to express and map their ideas into word sequence, and the word sequence is already cached in memory when human speaking or writing. In most case, the cached word sequence may be not a complete sentence but at least most part of it. On the other hand, for reading or listening, it is better to know both side context of a word when predicting the meaning of the word or define the grammar properties of the word. Therefore, it is not a good strategy to deal with word sequences in a natural language word by word in a certain order which has also been questioned by the success application of BiRNN in some NLP tasks.

Another limit of NNLM caused by model architecture is original from the monotonous architecture of ANN. In ANN, models are trained by updating weight matrixes and vectors which distribute among all nodes. Training will become much more difficult or even unfeasible when increasing the size of model or the variety of connections among nodes, but it is a much efficient way to enhance the performance of ANN. As is well known, ANN is designed by imitating biological neural system, but biological neural system does not share the same limit with ANN. In fact, the strong power of biological neural system is original from the enormous number of neurons and various connections among neurons, including gathering, scattering, lateral and recurrent connections \citep{nicholls_2014}. In biological neural system, the features of signals are detected by different receptors, and encoded by low-level central neural system (CNS) which is changeless. The encoded signals are integrated by high-level CNS. Inspired by this, an improvement scheme for the architecture of ANN is proposed, as illustrated in Figure \ref{fig:scheme}. The features of signal are extracted according to the knowledge in certain field, and every feature is encoded using changeless neural network with careful designed structure. Then, the encoded features are integrated using a trainable neural network which may share the same architecture as existing ones. Because the model for encoding does not need to be trained, the size of this model can be much huge and the structure can be very complexity. If all the parameters of encoding model are designed using binary, it is possible to implement this model using hardware and higher efficiency can be expected.

\begin{table}[!hbp]
\begin{center}
\begin{tabular}{c|c|c|c|c|c|c}
\toprule
\multirow{2}{*}{Model} & \multirow{2}{*}{$m$} & \multirow{2}{*}{$n_h$} & \multicolumn{2}{c|}{Training Data} & \multicolumn{2}{c}{PPL}\\
\cline{4-7}
 & & & Electronics & Books & Electronics & Books \\
\midrule
LSTM-NNLM & 100 & 200 & Yes & No & 146.62 & 353.16 \\
LSTM-NNLM & 100 & 200 & No & Yes & 280.91 & 246.35 \\
LSTM-NNLM & 100 & 200 & Yes & Yes & 147.28 & 302.92 \\
\bottomrule
\end{tabular}
\caption{Evaluating NNLM on data sets from different fields}
\label{tab:dynamic}
\end{center}
\end{table}

\subsection{Knowledge Representation}
The word "learn" appears frequently with NNLM, but what neural network language models learn from training data set is rarely analyzed carefully. The common statement about the knowledge learned by neural network language models is the probabilistic distribution of word sequences in a natural language. Strictly speaking, it is the probabilistic distribution of word sequences from a certain training data set in a natural language, rather than the general one. Hence, the neural network language model trained on data set from a certain field will perform well on data set from the same field, and neural network language model trained on a general data set may show worse performance when tested on data set from a special field. In order to verify this, one million words reviews on electronics and books were extracted from Amazon reviews \citep{he_2016,mcauley_2015} respectively as data sets from different fields, and 800000 words for training, 100000 words for validation, and the rest for test. In this experiment, two models were trained on training data from electronics reviews and books reviews respectively, and the other one was trained on both. Then, all three models were tested on the two test data sets.

The lowest perplexity on each test data set was gained by the model trained on corresponding training data set, instead of the model trained on both training data set (Table \ref{tab:usage}). The results show that the knowledge represented by a neural network language model is the probabilistic distribution of word sequences from training data set which varies from field to field. Except for the probabilistic distribution of word sequences, the feature vectors of words in vocabulary are also formed by neural network during training. Because of the classification function of neural network, the similarities between words can be observed using these feature vectors. However, the similarities between words are evaluated in a multiple dimensional space by feature vectors and it is hard to know which features of words are taken into account when these vectors are formed, which means words cannot be grouped according to any single feature by the feature vectors. In summary, the knowledge represented by neural network language model is the probabilistic distribution of word sequences from certain training data set and feature vectors for words in vocabulary formed in multiple dimensional space. Neither the knowledge of language itself, like grammar, nor the knowledge conveyed by a language can be gained from neural network language models. Therefore, NNLM can be a good choice for NLP tasks in some special fields where language understanding is not necessary. Language understanding cannot be achieved just with the probabilistic distribution of word sequences in a natural language, and new kind of knowledge representation should be raised for language understanding.

Since the training of neural network language model is really expensive, it is important for a well-trained neural network language model to keep learning during test or be improved on other training data set separately. However, the neural network language models built so far do not show this capacity. Lower perplexity can be obtained when the parameters of a trained neural network language model are tuned dynamically during test, as showed in Table \ref{tab:dynamic}, but this does not mean neural network language model can learn dynamically during test. ANN is just a numerical approximation method in nature, and it approximate the target function, the probabilistic distribution of word sequences for LM, by tuning parameters when trained on data set. The learned knowledge is saved as weight matrixes and vectors. When a trained neural network language model is expected to adaptive to new data set, it should be retrained on both previous training data set and new one. This is another limit of NNLM because of knowledge representation, i.e., neural network language models cannot learn dynamically from new data set. 

\begin{table}[!hbp]
\begin{center}
\begin{tabular}{c|c|c|c|c}
\toprule
Model & $m$ & $n_h$ & Dynamic & PPL \\
\midrule
LSTM-NNLM & 100 & 200 & No & 237.93 \\
LSTM-NNLM & 100 & 200 & Yes & 174.57 \\
\bottomrule
\end{tabular}
\caption{Examine the learning ability of neural network}
\label{tab:usage}
\end{center}
\end{table}

\section{Future Work}
Various architectures of neural network language models are described and a number of improvement techniques are evaluated in this paper, but there are still something more should be included, like gate recurrent unit (GRU) RNNLM, dropout strategy for addressing overfitting, character level neural network language model and ect. In addition, the experiments in this paper are all performed on Brown Corpus which is a small corpus, and different results may be obtained when the size of corpus becomes larger. Therefore, all the experiments in this paper should be repeated on a much larger corpus.

Several limits of NNLM has been explored, and, in order to achieve language understanding, these limits must be overcome. I have not come up with a complete solution yet but some ideas which will be explored further next. First, the architecture showed in Figure \ref{fig:scheme} can be used as a general improvement scheme for ANN, and I will try to figure out the structure of changeless neural network for encoder. What's more, word sequences are commonly taken as signals for LM, and it is easy to take linguistical properties of words or sentences as the features of signals. However, it maybe not a proper way to deal with natural languages. Natural languages are not natural but man-made, and linguistical knowledge are also created by human long after natural language appeared. Liguistical knowledge only covers the "right" word sequences in a natural language, but it is common to deal with "wrong" ones in real world. In nature, every natural language is a mechanism of linking voices or signs with objects, both concrete and abstract. Therefore, the proper way to deal with natural languages is to find the relations between special voices or signs and objects, and the features of voices or signs can be defined easier than a natural language itself. Every voice or sign can be encoded as a unique code, vector or matrix, according to its features, and the similarities among voices or signs are indeed can be recognized from their codes. It is really difficult to model the relation between voices or signs and objects at once, and this work should be split into several steps. The first step is to covert voice or sign into characters, i.e., speech recognition or image recognition, but it is achieved using the architecture described in Figure \ref{fig:scheme}.

\section{Conclusion}
In this paper, different architectures of neural network language models were described, and the results of comparative experiment suggest RNNLM and LSTM-RNNLM do not show any advantages over FNNLM on small corpus. The improvements over these models, including importance sampling, word classes, caching and BiRNN, were also introduced and evaluated separately, and some interesting findings were proposed which can help us have a better understanding of NNLM. 

Another significant contribution in this paper is the exploration on the limits of NNLM from the aspects of model architecture and knowledge representation. Although state of the art performance has been achieved using NNLM in various NLP tasks, the power of NNLM has been exaggerated all the time. The main idea of NNLM is to approximate the probabilistic distribution of word sequences in a natural language using ANN. NNLM can be successfully applied in some NLP tasks where the goal is to map input sequences into output sequences, like speech recognition, machine translation, tagging and ect. However, language understanding is another story. For language understanding, word sequences must be linked with any concrete or abstract objects in real world which cannot be achieved just with this probabilistic distribution. 

All nodes of neural network in a neural network language model have parameters needed to be tunning during training, so the training of the model will become very difficult or even impossible if the model's size is too large. However, an efficient way to enhance the performance of a neural network language model is to increase the size of model. One possible way to address this problem is to implement special functions, like encoding, using changeless neural network with special struture. Not only the size of the changeless neural network can be very large, but also the structure can be very complexity. The performance of NNLM, both perplexity and training time, is expected to be improved dramatically in this way.

\newpage

\vskip 0.2in
\bibliography{references}

\end{document}